# Residual Value Forecasting Using Asymmetric Cost Functions

Authors, Affiliations, and Postal address:

Korbinian Dress[1]

Stefan Lessmann[1]

Hans-Jörg von Mettenheim[2]

[1] *School of Business and Economics, Humboldt-University of Berlin, Unter-den-Linden 6, 10099 Berlin, Germany*

[2] *Institute of Information Systems, Leibniz Universität Hannover*

Email:

*k.dress@gmx.de*

*stefan.lessmann@hu-berlin.de*

*mettenheim@iwi.uni-hannover.de*

Corresponding Author:

*Stefan Lessmann [stefan.lessmann@hu-berlin.de]*

*Tel.: +49 (0)30 2093 5742*

*Fax: +49 (0)30 2093 5741*

# Residual Value Forecasting Using Asymmetric Cost Functions

**Abstract**

Leasing is a popular channel to market new cars. Pricing a leasing contract is complicated because the leasing rate embodies an expectation of the residual value of the car after contract expiration. To aid lessors in their pricing decisions, the paper develops resale price forecasting models. A peculiarity of the leasing business is that forecast errors entail different costs. Identifying effective ways to address this characteristic is the main objective of the paper. More specifically, the paper contributes to the literature through i) consolidating and integrating previous work in forecasting with asymmetric cost of error functions, ii) systematically evaluating previous approaches and comparing them to a new approach, and iii) demonstrating that forecasting with asymmetric cost of error functions enhances the quality of decision support in car leasing. For example, under the assumption that the costs of overestimating resale prices is twice that of the opposite error, incorporating corresponding cost asymmetry into forecast model development reduces decision costs by about eight percent, compared to a standard forecasting model. Higher asymmetry produces even larger improvements.

*Keywords: forecasting, artificial intelligence, ensemble learning, asymmetric cost of error*

# 1. Introduction

Forecasting Support Systems (FSS) aid managers in anticipating future developments to make informed decisions. The core of a FSS is a forecast model, the purpose of which is to produce operationally accurate predictions (e.g. Shmueli & Koppius, 2011).

The paper concentrates on decision support in car leasing and the prediction of resale prices in particular. Manufacturers that offer leasing contracts require a forecast of the car's residual value after contract expiration. The difference between the car's original list price and its residual value determines the leasing rate. In this respect, FSS are important to support pricing decisions. Generally, forecasts are not fully accurate, but will either over- or underestimate actual resale prices. From a managerial business perspective, there is little reason to believe that equivalent costs arise from these errors. Underestimating resale prices yields unexpected profits when selling the returned car in the second-hand market. However, it may carry opportunity costs. In retrospect, the lessor could have offered lower leasing rates to increase sales. Due to relatively high prices, some customers might have been deterred to sign a contract. On the other hand, overestimating resale prices implies that the lessor realizes lower profits from the overall contract or might face a loss. Therefore, the costs of forecast errors are asymmetric.

Granger (1969) was the first to suggest that real-world forecasting tasks are rarely characterized by quadratic (i.e., symmetric) error costs. Several others have echoed Granger's (1969) critic, developed asymmetric cost of error functions (ACEF), and demonstrated their potential to improve decision quality through empirical experimentation (e.g. Crone, 2010; Diebold & Mariano, 1995; Leitch & Tanner, 1991). Lessmann (2013) arrives at a similar conclusion for resale price modeling in the automotive industry.

The objective of this paper is threefold: (i) to consolidate and integrate previous work in forecasting using ACEF, (ii) to systematically compare previous and newly developed modeling approaches, and (iii) to demonstrate the potential of such approaches to improve pricing decision in the car leasing business. In pursuing these objectives, the paper makes the following contributions: First, it re-introduces several modeling strategies to account for unequal costs of positive and negative deviations including quantile regression, ANN, and machine learning algorithms. This provides novel insights concerning the relative merits of these approaches because they have, to the best of our knowledge, not been considered in a common decision support context. Second, the paper develops a conceptual framework to identify the stage in the forecasting process in which cost asymmetry is best accounted for. A comparison of alternative strategies demonstrates that considering asymmetry during model estimation is superior to an ex-post correction of predictions. In response, a novel ensemble modeling approach is proposed.

Third, to examine the interaction between the internal functioning of a forecast model and strategies to address asymmetric costs of errors, the paper systematically compares linear and non-linear forecasting models as well as individual and ensemble models. Previous forecasting benchmarks commonly analyze observed results along these dimensions. Yet, in the field of forecasting using ACEF, comparable evidence is missing.

From a large set of empirical experiments and subsequent sensitivity analysis, we identify a sizeable potential to improve the quality of decision support in the presence of asymmetric costs. For example, if the error of overestimating resale prices is weighted twice as much as the reverse error, the most efficient ACEF modeling strategy enhances decision quality by about eight percent compared to the best conventional prediction method. For higher degrees of asymmetry, the improvement is even larger.

The remainder of the paper is organized as follows: Section 2 provides some theoretical background and reviews related work. Section 3 elaborates on ACEF and strategies to account for asymmetric error costs. Section 4 introduces the experimental design, before results are presented and discussed in Section 5. Section 6 concludes the paper.

## 2. Theoretical background and related literature

### 2.1. Decision support in car leasing

Pricing is an important decision task in the car leasing business (Du et al., 2009). The price corresponds to the leasing rate. Given that it is the lessor's obligation to take back and remarket the returned car after expiration of the leasing contract, the (discounted) sum of payments implies an expectation of the car's residual value. Prediction of resale prices is thus important to the lessor to inform price setting and secure a profit from the lease.

Jerenz (2008) develops a decision support system to guide price setting in the used car business and evidences the potential of corresponding approaches to increase revenue. To model resale prices, Jerenz (2008) and others employ hedonic regression (e.g., Prado, 2009, and Erdem & Sentürk, 2009). The resulting explanatory models reveal the degree of which car characteristics (i.e., independent variables) affect resale prices, which is beneficial to gain insight into the formation of prices. However, from a practical point of view, a drawback of explanatory models is that they display lower accuracy than models that are deliberately constructed for prediction (Shmueli & Koppius, 2011). For pricing decision support, it is desirable to choose models that predict as accurately as possible (Du et al., 2009).

Lian et al. (2003) pioneer the use of advanced forecasting methods to predict resale prices. They use a neural network system on five different car models and let an evolutionary algorithm select the most important variables and the most suitable meta-parameters. Lessmann & Voß (2013) conduct a comprehensive comparison of 19 state-of-the-art prediction models in various scenarios and under different conditions to determine whether advanced methods improve forecast accuracy. Their results evidence that random forest regression is an especially effective method for resale price forecasting.

The previous studies focus on forecast accuracy. Yet, as explained in Section 1, overestimating the resale price of a returned car entails different costs than underestimating it. When the actual resale price of a leasing car is higher than estimated, the residual value is underestimated, which leads to unexpected profits at the time of resale. On the other hand, this is associated with opportunity costs in that a lower leasing rate could have been offered to increase sales. On the other hand, forecasting a residual value above the actual resale price (overestimating) leads to an unexpected decrease of profit – and maybe a loss – when remarketing the returned car. A risk-averse manager will thus consider overestimating to be the more severe error. Prospect theory suggests that decision makers weigh losses higher than gains even if they have the same magnitude and occur with equal probabilities (Kahneman & Tversky, 1979; Tversky & Kahneman, 1992). Therefore, a manager will consider overestimation to be the more severe forecast error.

2.2. Forecasting in the presence of asymmetric costs

Forecasting methods are manifold and include linear, nonlinear and ensemble methods. A common characteristic across different approaches is that forecast errors are evaluated by quadratic and symmetric cost functions. Referring to the mismatch of such cost functions and those occurring in real-world business scenarios, Granger (1969) criticizes this approach and demonstrates that forecasts based on quadratic error functions for general cost functions[1] do not lead to optimal estimators. Therefore, the estimation and assessment of a forecast should focus on actual economic costs (e.g., Granger & Newbold, 1986; Diebold & Mariano, 1995; Christoffersen & Diebold, 1997; Leitch & Tanner, 1991). More generally, the assessment of a decision support system should focus on the system's ability to improve decision quality and business performance (Lilien et al., 2004). Again, this suggests evaluation criteria such as profits and costs (Bharadwaj, 2000) and questions the prevailing use of symmetric cost functions in forecasting.

---

[1] Granger uses the term *cost function* rather than the term *error* or *loss function*. However, in the current context, these are words for the same thing and hence will be used interchangeably in the following.

Some studies employ asymmetric cost functions in real-world forecasting applications.[2] In the car leasing context considered here, Lessmann (2013) reports preliminary findings, on which this study further elaborates. Examples in other domains include Tian (2009), who forecasts Australian unemployment rates considering a quadratic asymmetric loss function. She argues that an unemployment rate below the target causes much lower costs than overestimating unemployment rates. Exploiting the ability to model conditional quantiles rather than the conditional mean, Berk (2011) uses quantile regression to develop forecasts for crime prevention. Specifically, if the number of robberies in an area is overestimated, the area will be observed. In the opposite case, it will receive insufficient surveillance, which could lead to a fatal increase in crime (Berk, 2011). Crone (2002, 2010) trains artificial neural networks with asymmetric cost functions to optimize the replenishment of automatic vending machines. In this context, cost asymmetry results from the fact that a surplus of goods causes lower expenses compared to a stockout. This is because surplus materials bind capital, occupy sales floor and may cause additional costs related to risk of damage, burglary and deterioration. Shortfalls, on the other hand, suffer from more severe consequences. They cause direct profit loss and lead to reduced customer satisfaction in the long run (Crone, 2010). Similar considerations apply to almost every supply chain and re-emphasize the importance and prevalence of cost asymmetry in real-world applications; known for a long time (Granger, 1969) but no less relevant today.

## 3. Methodology

In general, the forecasting process splits into multiple steps (Figure 1). Once a prepared set of data is available, the first step is model estimation, which uses past data to determine the parameters of the forecasting model. For example, linear regression is the prevailing approach to estimate car resale prices. Subsequently, the model is used to generate forecasts for novel data. The forecasts of different models may be combined, which is a common way to increase accuracy (Timmermann, 2006). The combination, called an ensemble, requires to weigh the individual models. These weights are determined from past data and facilitate calculating the composite forecast for novel data.

---

[2] Note that there are several studies that consider cost asymmetry when predicting discrete target variables, for example in the field of cost-sensitive learning (e.g., Bleich, 2015). We do not consider these approaches here since the focus of the paper is on forecasting models for continuous targets.

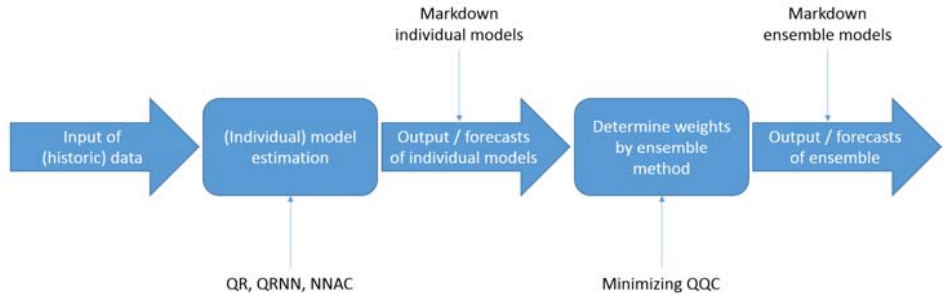

*Figure 1: Illustration of the forecasting process including the steps where cost asymmetry can be considered.*

Figure 1 illustrates how the approaches used in previous literature address asymmetric error costs at different process stages. Quantile regression (QR), quantile regression neural networks (QRNN) and neural networks with asymmetric cost functions (NNAC) account for cost asymmetry when estimating the forecast model, for example through using an ACEF for model fitting (e.g., Crone, 2010). The markdown procedure operates differently and suggests an ex post correction of individual or ensemble model forecasts. A third option arises in ensemble modeling. In particular, the weights of individual models in an ensemble can be chosen such that an ACEF is minimized (e.g., Lessmann, 2013). Subsequent chapters give a formal description of ACEFs and the above options in the context of resale price forecasting.

3.1. Asymmetric cost of error functions

Error measures are a common way to evaluate forecast performance in a quantitative manner. They are designed on the basis of loss (or cost) functions, which penalize deviations between forecasts and actuals. The most common loss function is quadratic loss:

$$C(e_i) = e_i^2, \text{ with } e_i = (y_i - \hat{y}_i), \quad (1)$$

where $\hat{y}_i$ represents the model-based forecast for case *i*. The corresponding error measurement is the well-known mean squared-error (MSE).

$$MSE = \frac{1}{n}\sum_{i=1}^{n} e_i^2. \quad (2)$$

Quadratic loss implies the structure of error costs to be such that they increase quadratically with the magnitude of deviations between forecasts and actuals, whereby the direction of the deviation is irrelevant. As explained above, this is implausible in car leasing (Lessmann, 2013).

Formally, asymmetric cost functions include the linear-linear cost (LLC), quadratic-quadratic cost (QQC), and linear-exponential cost (LEC) (e.g., Christoffersen & Diebold, 1997; Crone, 2010):

$$LLC\ (e_i) = \begin{cases} a|e_i|, if\ e_i > 0 \\ b|e_i|, if\ e_i \leq 0 \end{cases}; \quad QQC\ (e_i) = \begin{cases} ae_i^2, if\ e_i > 0 \\ be_i^2, if\ e_i \leq 0 \end{cases};$$

(3)

$$LEC\ (e_i) = b[exp(ae_i) - ae_i - 1]$$

The parameters $a$ and $b$ facilitate adjusting the ACEF to the empirical cost situation in that they determine the severity of a particular error type. For example, setting $a = b$, QQC reduces to MSE. Setting $a < b$ indicates that overestimating resale prices is more costly than underestimation; with analogous implications for $a > b$. Note that the above ACEF fulfil the requirements for generalized cost functions (Granger, 1999; Diebold, 2001). First, a prediction error of zero ($C(0) = 0$) incurs no loss. Second, the loss remains positive for all – positive and negative – forecast deviations ($C(e_i) > 0$ for $e_i \neq 0$). Last, $C(e_i)$ is monotonically increasing in $|e_i|$. Table 1 summarizes the different error measures and costs associated with (3), whereas Figure 2 depicts how they relate forecast error size and direction to error costs.

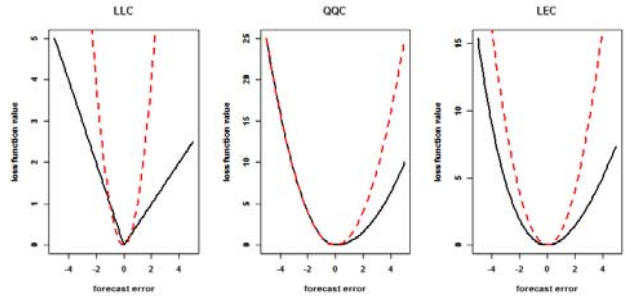

*Figure 2: Asymmetric cost functions (solid black line) versus squared error loss (red dotted line). The values for a and b are 0.5 and 1 for LLC, 0.4 and 1 for QQC, and 0.22 and 17 for LEC.*

*Table 1: Summary of the asymmetric cost functions and the corresponding error measures.*

|  | linear-linear cost function | quadratic-quadratic cost function | linear-exponential cost function |
|---|---|---|---|
| simple error | $LLC(e_i) = \begin{cases} a\|e_i\|, if\ e_i > 0 \\ b\|e_i\|, if\ e_i \leq 0 \end{cases}$ | $QQC(e_i) = \begin{cases} ae_i^2, if\ e_i > 0 \\ be_i^2, if\ e_i \leq 0 \end{cases}$ | $LEC(e_i) = b[exp(ae_i) - ae_i - 1]$ |
| mean error measure | $MLLC(e_i) = \frac{1}{n}\sum_{i=1}^{n} LLC(e_i)$ | $MQQC(e_i) = \frac{1}{n}\sum_{i=1}^{n} QQC(e_i)$ | $MLEC(e_i) = \frac{1}{n}\sum_{i=1}^{n} LEC(e_i)$ |

ACEFs remain an approximation of actual costs in resale price forecasting and car leasing. For example, they do not calculate a monetary cost. Also, the monetary loss of a leasing contract cannot be arbitrarily large, which none of the above ACEF takes into account. However, they represent the degree to which a forecasting model supports pricing decisions much better than symmetric error functions. Therefore, examining the degree to which using ACEF during forecast model development improves over standard practices provides valuable insights related to model-based decision support in car learning. Corresponding approaches are introduced in the following sections.

3.2. Linear methods that account for asymmetric costs

The simple markdown, also called "*Granger's simpler procedure*" (Weiss, 1996), accounts for asymmetric error cost through an ex-post adaption of forecasts, which have been developed based on a standard cost function (Tian, 2009). A slightly revised version of this approach is used here. Instead of adding a constant absolute bias term, forecasts are adapted through adding a constant percentage. More specifically, the paper employs standard prediction models, which do not account for cost asymmetry, and calculates an optimal percentage to change forecasts ex post so as to minimize the loss from an ACEF.

For example, we might estimate the parameters of a linear regression model using OLS, calculate forecasts using the estimated parameters, and subsequently adapt the forecasts $\hat{f}(X) = \hat{y}$ through subtracting a constant percentage $md$:

$$\hat{y}_i^{md} = \hat{y}_i\,(1 - md). \tag{4}$$

We obtain the estimate for $md$ through minimizing the forecasting loss occurring from the deviation between the markdown prediction and the actual realization of the dependent variable measured by an ACRF, Considering the QQC function as example, this equates to:

$$\min_{md} \sum_{i=1}^{n} QQC[y_i - \hat{y}_i(1 - md)]. \tag{5}$$

One may interpret $md$ as a depreciation rate or markdown on the predicted resale price of a returned car due to the higher severity of overestimating. We will apply this procedure to the case of a linear model (MBL), a nonlinear model (MBNL) and an ensemble (ES_md), using the best[3] prediction model of each class. Conceptually, the simple markdown procedure has same advantages. First, it is easy to interpret. The markdown represents a fixed percentage that is subtracted from the forecast to account for decision makers' preferences and cost asymmetries between positive and negative deviations. Second, it is easy to implement and can be applied to any forecasting method.

A second linear method that accounts for asymmetric costs is quantile regression (QR). Similarly to estimating a conditional mean through minimizing the sum of squared residuals, it is possible to determine a conditional quantile as the solution to the problem of minimizing the weighted sum of absolute residuals. In particular, the $\tau$-th quantile in a sample $\{y_1, \ldots, y_n\}$ can be estimated as follows (Koenker & Bassett, 1978; Koenker & Hallock, 2001):

$$\min_{\xi} \sum_{i=1}^{n} \rho_\tau(y - \xi) = \sum_{i=1}^{n} [\tau I(y_i > \xi) + (1-\tau)I(y_i \leq \xi)]|y_i - \xi|, \quad (6)$$

where $I(.)$ is an indicator function, which is equal to 1 if the event is true and 0 otherwise and $\rho_\tau$ is a loss function. The conditional quantiles, which represent the actual quantile regression, are determined by a function of independent variables, such that the predictions fall on a hyperplane in space defined by the response and explanatory variables. To obtain the estimate of the conditional quantile regression, $x_i'\beta$ simply replaces $\xi$ in (6):

$$\min_{\beta} \sum_{i=1}^{n} \rho_\tau(y_i - x_i'\beta). \quad (7)$$

---

[3] Best means most accurate in the case of a standard symmetric loss function.

The optimization problem is comparable to the least squares regression, but instead of minimizing the sum of squared residuals, it minimizes the weighted sum of absolute values of the residuals. The regression fit $x_i'\beta$ describes the $\tau$-th quantile of the response variable $y_i$ (i.e., resale price of car $i$) given the characteristics $x_i$ (i.e., leasing contract duration, mileage, etc.)). In contrast to OLS, it is possible to weigh positive and negative residuals differently.

Equation 7 is expressed in a loss function that directly includes a deviation score. The deviation of the potential estimate of the $\tau$-th quantile $\xi$ from the actual value $y_i$ is defined by $d_i = y_i - \xi$, with $d_i$ denoting a deviation score. Instead of the mean, it is computed for a quantile. The resulting quantile loss function is:

$$L(d_i) = \sum_{i=1}^{n} d_i \left(\tau - I(d_i < 1)\right). \quad (8)$$

The loss function is the deviation score $d_i$ multiplied by either $\tau$ or $1 - \tau$, summed over all cases in the data set. The resulting $\tau$-th quantile is the value of $Y$ that yields the lowest loss of quantile loss function.

For $\tau = 0.5$ the corresponding quantile loss functions of an exemplary deviation score of $d_i = 4$ and $d_i = -4$ are both 2. However, when $\tau = 0.25$, the deviation score of $d_i = -4$ yields a loss of 1, while the loss of $d_i = -4$ is 3. This demonstrates the influence of the deviation score and how it facilitates an asymmetric weighting of forecast errors. The weights are determined by the choice of the appropriate $\tau$. The weighting relation from negative to positive deviations is characterized by $\frac{(1-\tau)}{\tau}$. In general, $\tau < 0.5$ renders negative deviations more costly. The opposite holds for $\tau > 0.5$. This illustrates that there is a direct relationship between the loss function and the quantile to be forecast. Hence, $\tau$ determines the asymmetry of the loss function. Moreover, it is possible to replicate the weighting of the $LLC(a,b)$ by calculating the corresponding $\tau$ by $\tau = \frac{a}{1+a}$, where $b = 1$.

Using QR to account for asymmetric costs has four advantages: First, it is a method that is well-grounded in statistical theory and has been successfully applied in several studies to compute conditional quantiles (e.g., Fitzenberger et al., 2002). However, the possibility of treating negative and positive residuals differently is rarely exploited. Second, QR is a parametric linear method. Model coefficients can be interpreted in the same way as in OLS regression. In car resale price forecasting, this provides value information to decision makers and shares similarities with previous work on hedonic regression of used car prices (e.g., Erdem & Sentürk, 2009). Third, efficient linear programming methods are available to solve (7). Hence, using QR is neither time- nor resource-intensive. Last, unlike the markdown procedure, which is an ex post approach, QR introduces asymmetric error costs ex ante and thus at an earlier modeling stage than the markdown procedure. An earlier consideration of application characteristics might increase the (business) value of the forecasting method.

3.3. Artificial neural networks considering asymmetric costs

Artificial neural networks (NN) are a family of nonparametric, nonlinear forecasting methods. The most common single hidden layer feedforward NN (Zhang et al., 1998) has $m$ inputs (i.e. explanatory variables $x_{ij}$), $k$ nodes in the hidden layer, and one output neuron, which models the dependent variable $y_i$. The formal expression reads:

$$f(x_i, v, w) = g_2\left(\sum_{e=0}^{k} v_e g_1\left(\sum_{j=1}^{m} w_{je} x_{ij}\right)\right), \qquad (9)$$

where $g_1(.)$ and $g_2(.)$ denote activation functions that enable the NN to model nonlinear relationships, and $w_{je}$ and $v_e$ represent connection weights (i.e. the model parameters, which will be estimated form data).

White (1992) provides theoretical support for the integration of quantile regression within a NN framework for the estimation of nonlinear quantile models. Similar to fitting a linear quantile function using equation (7), a quantile regression neural network (QRNN) estimation, $f(x_i, v, w)$, of the $\tau$-th quantile includes a comparable loss function. The solution can be found by optimizing the following minimization:

$$\min_{v,w}\left(\sum_i \rho_\tau(y_i - f(x_i, v, w)) + \lambda_1 \sum_{e,j} w_{je}^2 + \lambda_2 \sum_e v_e^2\right). \qquad (10)$$

The regularization parameters $\lambda_1$ and $\lambda_2$ avoid model overfitting through penalizing the complexity of the network (Bishop, 1997). These two combined with the third meta-parameter, namely the number of hidden nodes $k$, are determined by cross-validation (e.g., Donaldson & Kamstra, 1996).

Since the QRNN equation includes an equivalent loss function as QR, asymmetric costs of over- and underestimation are regarded in the same manner. The quantile $\tau$ has to be set such that it mimics the asymmetry between positive and negative residuals, which allows QRNN to capture characteristics of car resale price forecasting better than a NN trained with a symmetric cost function (e.g., Taylor, 2000). Another advantage of QRNN is that it is able to model nonlinear relationships, which often increases forecast accuracy (e.g., Crone, 2010). However, the computational cost associated with solving (10) to estimate a QRNN model is higher than in the case of, e.g., QR or ordinary NN.

An alternative strategy to adapt NN learning for forecasting tasks that display asymmetric error costs is to directly use an ACEFs for network training (Crone, 2002, 2010). We apply this approach for car resale price forecasting. In particular, we tune the connection weights in a three-layer feedforward neural network using QQC. Given that NN learning algorithms such as backpropagation employ gradient information, it is important to implement asymmetric error weighting in a differentiable way, which we achieve through the following sigmoidal approximation of QQC:

$$QQC_i^{appx} = e_i^2 \left(\frac{1}{1 + exp(99e_i)}\right)(b - a) + a. \tag{11}$$

The idea is to employ the logistic function $sig(e_i) = \frac{1}{1+\exp(-e_i)}$ as a means to model the step function in QQC, which penalizes positive and negative residuals differently. Adding a constant of 99 increases the steepness of the logistic function leading to a better approximation of the step function. Switching the sign of the exponent ensures that the logistic decreases with decreasing $e$.

3.4. Ensemble selection based on asymmetric error minimization

Much research has evidenced the merit of combining the forecasts of multiple models to increase accuracy (e.g., Timmermann, 2006). Ensemble selection (ES) is a forecast combination approach that first creates a library of base models and subsequently pools the forecasts of a subset of the base models by means of a weighted average (Caruana et al., 2004, 2006). ES has been shown to work well in several applications including car resale price forecasting using ACEF (Lessmann, 2013). The paper further elaborates on the methodology of Lessmann (2013) in that we augment the model library with ACEF-based forecasts. We find this to improve forecast accuracy substantially (see below).

The first step of ES comprises developing a set of alternative forecasting models. To that end, we employ different forecasting methods including linear, nonlinear and (homogeneous) ensemble methods. We also consider multiple meta-parameter settings for individual forecasting methods. Table 2 summarizes the model library. A detailed description of the forecasting methods is beyond the scope of the paper and available in, e.g., Hastie et al. (2009).

The approach to select and combine a subset of models from the model library has been proposed by Caruana et al. (2004, 2006). They use a stepwise forward selection heuristic to form the ensemble. The first candidate ensemble corresponds to the base model with highest accuracy. Then, all candidate ensemble including this model and one other model from the library are evaluated. Provided that the most accuracy ensemble with two members improves upon the best individual base model, ensemble selection continues; and terminates otherwise. Table 3 exemplifies the algorithm with a hypothetical library of three base models $M_1, M_2$ and $M_3$, their forecasts for three observations (e.g., cars resale prices) and the corresponding actual values. Each of the four panels corresponds to one iteration. The rightmost column depicts the ensemble composition of the corresponding iteration. Considering the MSE as performance measure, base model $M_3$ predicts most accurately. Therefore, the ensemble after the first iteration corresponds to $M_3$. The second panel shows the MSE for all possible combinations of the present ensemble (i.e., $M_3$) and one other base model. Considering the example of $M_1$, the forecasts shown in Panel 2 are calculated as a simple average of the forecasts of $M_1$ and $M_3$. Panel 2 reveals that the combination of $M_2$ and $M_3$ has lower error than $M_3$. Thus, $M_2$ is added to the ensemble. In summary, the ensemble after the second iteration consists of $M_2$ and $M_3$ and the composite forecast is the simple average of their predictions. Panel 3 depicts the forecasts of all ensemble with size three and including $M_2$ and $M_3$ and illustrates how a second addition of $M_3$ further increases the accuracy of the composite forecast. This shows that ES effectively computes a weighted average of base model forecasts (Caruana et al., 2004). The best ensemble in the fourth iteration (Panel 4) is obtained when further increasing the weight of $M_3$ in the composite forecast through adding it a third time to the ensemble. This addition, however, does not increase forecast accuracy compared to the ensemble of the previous iteration ($MSE_{Panel4} = 33.52 > MSE_{Panel3} = 32.80$). Therefore, the ES terminates after four iteration. The final ensemble consists of $M_3$, $M_2$ and $M_3$. The ensemble forecast is equivalent to a weighted average with weights of $\frac{2}{3}$ for $M_3$, $\frac{1}{3}$ for $M_2$ and 0 for $M_1$.

Table 2: Overview of regression methods included in the base model library.

| Regression method | # of models | Meta-parameters | Type | Pattern |
|---|---|---|---|---|
| Multivariate linear regression | 1 | - | individual | linear |
| Stepwise linear regression | 2 | forward, backward | individual | linear |
| Ridge regression | $3 + 3 \cdot 7 = 24$ | # of principal components, scaling method, lambda | individual | linear |
| Lasso | 16 | steps on the path | individual | linear |
| Mars | $3 \cdot 3 \cdot 3 = 27$ | penalty, threshold, degree | individual | linear |
| Neural network | $5 \cdot 8 = 40$ | # of hidden neurons, threshold | individual | nonlinear |
| Support vector regression | $2 \cdot 4 \cdot 5 = 40$ | cost, tolerance, epsilon | individual | nonlinear |
| Regression tree | $5 \cdot 5 = 25$ | complexity, min # of observations in a node | individual | nonlinear |
| k nearest neighbor | $3 \cdot 10 = 30$ | type of algorithm, # of neighbors | individual | nonlinear |
| Bagged regression tree | 7 | # of bags | ensemble | nonlinear |
| Bagged neural network | 7 | # of bags | ensemble | nonlinear |
| Boosted regression tree | 5 | maximal depth | ensemble | nonlinear |
| Random forest | $8 \cdot 5 = 40$ | # of trees, # of variables | ensemble | nonlinear |
| **Total** | **264** | | | |

Table 3: Example of ensemble selection with a hypothetical library of three base models.

| | | $y$ | $M_1$ | $M_2$ | $M_3$ | Ensemble |
|---|---|---|---|---|---|---|
| Panel 1 | $car_1$ | 53.66 | 62.90 | 65.63 | 61.26 | $M_3$ |
| | $car_2$ | 45.36 | 35.76 | 47.91 | 38.92 | |
| | $car_3$ | 67.07 | 66.90 | 65.63 | 64.50 | |
| | MSE | | 59.19 | 50.62 | **35.29** | |
| Panel 2 | $car_1$ | 53.66 | 62.08 | 63.44 | 61.26 | $\frac{M_3+M_2}{2}$ |
| | $car_2$ | 45.36 | 37.34 | 43.41 | 38.92 | |
| | $car_3$ | 67.07 | 65.70 | 65.06 | 64.50 | |
| | MSE | | 45.70 | **34.53** | 35.29 | |
| Panel 3 | $car_1$ | 53.66 | 63.17 | 64.54 | 62.35 | $\frac{M_3+M_2+M_3}{3}$ |
| | $car_2$ | 45.36 | 39.59 | 45.66 | 41.17 | |
| | $car_3$ | 67.07 | 65.98 | 65.35 | 64.78 | |
| | MSE | | 41.67 | 40.47 | **32.80** | |
| Panel 4 | $car_1$ | 53.66 | 62.62 | 63.99 | 61.80 | $\frac{M_3+M_2+M_3+M_3}{4}$ |
| | $car_2$ | 45.36 | 38.46 | 44.54 | 40.04 | |
| | $car_3$ | 67.07 | 65.84 | 65.20 | 64.64 | |
| | MSE | | 43.16 | 36.97 | **33.52** | |

ES leaves a forecaster much freedom how to implement the assessment of candidate ensemble. Table 3 illustrates ES using the MSE as performance criterion. However, arbitrary indicators of forecast accuracy can be incorporated. More generally, Lessmann (2013) proposes to reserve the selection stage for optimizing business performance. Empirical results related to car resale price forecasting and ACEFs illustrate the potential of this approach. Therefore, we follow Lessmann (2013) and use the QQC function to guide ES. A conceptual advantage of this approach is that it integrates established forecasting principles (in the form of the base models in the library) and characteristics of the actual business application (represented by asymmetric error costs in car resale price forecasting).

## 4. Experimental Design

The experimental design aims at investigating the potential of incorporating error cost asymmetry into forecast development and at identifying the most suitable strategy to do so. To that end, we perform an empirical comparison of the forecasting methods discussed above using a real-world data set associated with pricing leasing contracts. To structure the analysis, we develop four research questions, which the empirical study will answer.

### 4.1. Research Questions

To guide readers through the empirical evaluation, we develop a set of four research questions (RQ). In this paper, the goal of forecast model development is to support pricing decisions in car leasing. In general, the value of a decision support model depends on the degree to which it increases decision quality. Given that pricing decisions in car leasing display asymmetric error costs, we suggest to approximate business objectives and accordingly decision quality through an ACEF. Then, a first RQ to be answered is whether the development of a forecasting model should take cost asymmetry into account. This leads to RQ1:

**RQ1: Are forecasting methods that account for cost asymmetry more effective than standard methods that use symmetric loss functions?**

Cost asymmetry can be considered at different modeling steps (see Figure 1). Broadly, available options split into an ex ante and an ex post approach. The latter bias forecasts but do not affect forecast model development. The former embody a more direct approach and alter the way in which models are developed. These opportunities together with the fact that an ex post correction is easier to implement motivate the second RQ:

**RQ2: Is it necessary to account for cost asymmetry during forecast model development building or is an ex post correction sufficient?**

In car resale price forecasting, the relationship between the dependent variables and the covariates is likely to be nonlinear (Desai & Purohit, 1998). For example, used cars lose much of their original value in the beginning of the usage period, which suggests a dampening effect. Moreover, introductions of new car models or major redesigns affect the resale prices of older models substantially and cause discontinuities in the relationship between resale prices and age (e.g., Purohit, 1992). On the other hand, estimating nonlinear models is complex, which might erode their advantage over simpler linear models. In general, the relative merits of nonlinear versus linear forecasting methods have received much attention in the literature. However, we are not aware of a systematic comparison of linear versus nonlinear modeling strategies to address asymmetric error costs. Therefore, to clarify the relative difference in forecasting performance in the focal application setting, we examine:

**QR3: To which extent do nonlinear forecasting methods improve performance over linear methods when the costs of forecast errors are asymmetric?**

The last RQ also concerns the merit of alternative modeling strategies. While several previous studies have established the efficacy of forecast combination and that combined models typically predict more accurately than individual models, corresponding evidence for the field of forecasting using ACEF is lacking. To the best of our knowledge, only the study of Lessmann (2013) provides some preliminary results related to this matter. Given the sparsity of prior work in the field, we examine:

**RQ4: Is it effective to pool forecasts when the costs of forecast errors are asymmetric?**

The research questions shed light on the relative effectiveness of alternative forecasting approaches in car leasing. We consider effectiveness to be a measure of business performance. The next section discusses how we approximate business performance.

### 4.2. Approximation of business objectives in car leasing

A cost of error function should represent the actual business objective (Granger, 1969). However, identifying 'the business goal' in a leasing context is not straightforward. Objectives depend on individual decision makers and vary across companies and market segments (Pierce, 2012). To avoid overspecialization and ensure relevance to a wide range of lessors, it is necessary to abstract the problem. Assuming that decision makers aim at minimizing the costs of suboptimal pricing decision provides such abstraction. Although lessors may likewise pursue growth targets or other business objectives, improved pricing decisions will, ceteris paribus, increase company profits (e.g., Marn et al., 2003). In this sense, minimizing the costs of erroneous pricing decisions appears a justifiable proxy for actual business objectives in car leasing. Under this assumption, it follows that ACEF are a suitable approximation of error costs. To see this, recall that prices (i.e., leasing rates) depend on resale price forecasts and that the costs of mispricing depend on the type of forecast error. This is exactly the cost structure that ACEFs embody. Assigning different weights to positive and negative forecast errors, they account for an important characteristic of the leasing business and are better aligned with actual business objectives than symmetric cost functions such as MSE. Therefore, the paper uses ACEFs not only to adapt forecast model development but also to measure the effectiveness of model-based decision support and thus as an approximation of business objectives in car leasing.

In particular, the analysis employs the QQC function. Compared to other ACEFs (see Section 3.1), QQC is relatively similar to the squared error function, which offers two advantages for this study. First, given that many forecasting methods rely on the (symmetric) squared error cost function, selecting QQC reduces the influence of factors other than cost asymmetry on observed results. Second, when assessing forecast models that originate from QQC minimization, similarity with MSE ensures that ordinary least-squares-based methods give a challenging benchmark.

### 4.3. Data description

A leading German car manufacturer, who prefers to remain anonymous, has provided a data set of 150,000 returned and resold leasing cars. The vehicles belong to the middle and premium segment of the passenger car market. The data comprises resales from the manufacturer during the period September 2005 to March 2015, whereby more than 95% of the sales occurred after Jan. 2011. The response variable is given by the ratio of a car's actual resale price over its original list price. It is common practice to represent residual values in this format (e.g., Prado, 2009; Lessmann & Voß, 2013). In addition, relative prices serve the data donor in that absolute prices and residual values remain confidential.

The independent variables characterize each transaction and capture: the age of the car, the duration of the leasing contract, special lacquers, a summary of customization characteristics, mileage, cubic capacity, horsepower, a dummy for four-wheel-drive, the type of motor fuel and the type of gear shift. Overall, the data set includes fifteen independent variable, which follow from encoding categorical variables through dummies.

To simulate a real-world application of alternative forecasting methods, where outcomes are unknown at the time of model building (e.g., Collopy et al., 1994), we reserve 30% of the data as a hold-out test set. We evaluate the performance of forecasting models on the test set using the mean error measure of the QQC function (MQQC). The 70% training set is further partitioned into an actual training set (ATS) and a validation set using a ratio of 4/7 and 3/7, respectively. We estimate forecast models with different meta-parameters (see Table 2) on the ATS. The purpose of the validation set is to i) determine the best models of different methods (e.g., the best NN model) and the best overall base model, which we use a benchmarks, to ii) implement base model selection within ES, and to iii) determine the optimal percentage of markdown $md$ of the markdown procedure. These modeling steps entail an assessment of forecasting methods and thus require auxiliary out-of-sample data. After taking the above modeling decision, we combine the ATS and the validation set and estimate the final forecasting models that are subsequently compared to each other using the test set.

## 5. Empirical results and interpretation

The empirical study aims at answering the four research questions and, more generally, to demonstrate the merit of considering asymmetric costs in car resale price forecasting. Another objective of the study is to identify the most effective approach to capture asymmetric costs.

Given that the costs of errors are often imprecise and depend on decision makers' preferences, we consider ten levels of asymmetry. The levels are denoted by the value of the parameter $a$. Assuming error weights to be positive, scaling QQC by a factor $1/b$ does not alter the cost minimal solution. Therefore, it is sufficient to study the behavior of QQC and corresponding forecasting models under changes of $a$ with fixed $b=1$. We choose $a$ because it is associated with positive residuals and thus underestimating resale prices. Recall that we consider this error to be less costly compared to overestimating resale prices. For example, a setting of $a = 0.1$ suggests that the error costs following from underestimating resale prices are ten times lower than those associated with overestimating. Likewise, $a = 0.2$ suggests that overestimating resale prices is five times more costly, while QQC equals MSE for $a = 1$. Hence, the lower $a$, the higher the degree of asymmetry.

### 5.1. Linear vs. Nonlinear Methods

A first set of experiments compares linear and nonlinear forecasting methods with and without ACEFs. More specifically, we consider ordinary least-squares linear regression (Lin. Reg.) as baseline and compare it to two linear and three nonlinear methods that use QQC for forecast model development. Quantile regression (QR) and the markdown of the best linear model (MBL) from the model library (see Table 2) are the two linear contestants. The nonlinear models include the quantile regression neural network (QRNN), a neural network trained with QQC (NNAC) and the markdown of the best nonlinear model (MBNL) from the model library (see Table 2). Table 4 reports the performance of these models in terms of MQQC across different settings of $a$. Smaller values indicate lower forecast error and thus better performance. Table 4 provides answers to research question 1 to 3. Its last three columns identify the models that enter corresponding comparisons.

*Table 4: Linear vs. Nonlinear remedy methods, evaluated by MQQC*

| Linear vs Nonlinear (MQQC) | | | | | | | | | | | RQ | | |
|---|---|---|---|---|---|---|---|---|---|---|---|---|---|
| Method | a=0.1 | a=0.2 | a=0.3 | a=0.4 | a=0.5 | a=0.6 | a=0.7 | a=0.8 | a=0.9 | a=1.0 | 1 | 2 | 3 |
| Lin. Reg. | 55.09 | 58.07 | 61.05 | 64.03 | 67.01 | 69.99 | 72.97 | 75.95 | 78.93 | 81.91 | | | |
| QR | 22.40 | 32.72 | 40.96 | 48.12 | 54.54 | 60.46 | 65.98 | 71.18 | 76.15 | 80.86 | | | |

| QRNN | 19.96 | 28.92 | 35.66 | 41.86 | 47.05 | 50.68 | 54.85 | 59.45 | 62.26 | 66.22 | | |
| NNAC | 21.18 | 29.70 | 36.30 | 41.98 | 47.28 | 51.51 | 56.66 | 60.83 | 63.62 | 66.11 | | |
| MBL | 25.36 | 36.48 | 44.84 | 51.82 | 57.94 | 63.47 | 68.55 | 73.29 | 77.75 | 81.56 | | |
| MBNL | 22.11 | 30.63 | 36.89 | 42.05 | 46.54 | 50.57 | 54.27 | 57.69 | 60.90 | 63.56 | | |

With respect to RQ1, Table 4 demonstrates that QR performs better than ordinary linear regression for all levels of asymmetry. It is remarkable that this includes the special case $a = 1$. A possible explanation is that QR regards the absolute error and is thus more robust towards outliers than Lin. Reg. This evidences that QR might be a suitable alternative to Lin. Reg. even if error costs are symmetric. More generally, we find linear regression to produce higher error than any other method. Consequently, accounting for cost asymmetry reduces the (asymmetric) costs of forecast errors and improves the performance of the forecasting model. This answers RQ1 and suggests that forecasting models to inform pricing decisions in car leasing should address the asymmetric costs that forecast errors carry.

The next research question (RQ2) concerns the step in the model building process at which cost asymmetry is best addressed. Table 4 includes two strategies. QR and QRNN consider asymmetry during model building. MBL and MBNL estimate a model using a symmetric cost function and account for asymmetry ex-post. The results of Table 4 offer clear advice for linear methods: QR is superior to MBL for all levels of asymmetry. Among the nonlinear methods, QRNN gives lower costs than MBNL for high degrees of asymmetry ($a = 0.1, ..., 0.4$). For $a \geq 0.5$, MBNL provides better results. In appraising the inconsistency in the MBNL vs. QRNN comparison, it is important to note the influence of the factor forecasting methods. MBNL corresponds to an ex post adaption (markdown) of the forecasts of the best nonlinear model from the model library (Table 2). This model is based on support vector regression. In terms of MSE, support vector regression predicts car resale prices more accurately than neural networks. Thus, better performance of MNBL compared to QRNN at lower levels of cost asymmetry may partially originate from superiority of support vector regression over neural networks. In summary, Table 4 provides some evidence in favor of an ex ante approach which accounts for asymmetry of error costs during model development, especially if the costs associated with positive and negative residuals differ substantially.

Relatively better performance of MBNL over QRNN and NNAC at lower levels of asymmetry hints at the importance of the expressive power of a forecasting method when forecasting with ACEFs. Being more flexible than linear methods, where QRNN performs consistently better than QR, nonlinear techniques methods are able to cope with mild degrees of asymmetry, provided that a suitable ex post correction of forecasts is applied. The purpose of the third research question is to examine the behavior of linear versus nonlinear forecasting methods. Table 4 reveals that nonlinear methods achieve lower MQQC than linear contestants across all levels of asymmetry. In particular, Table 4 suggests that MBL produces higher error costs than alternative methods. QR displays the second to worst performance and gives higher error costs than NNAC, MBNL and QRNN. Among these three, we find NNAC to perform consistently inferior to QRNN, which suggest the latter to be a more suitable approach to develop nonlinear forecasting models when facing asymmetric error costs. A comparison of QRNN to MBNL shows mixed results (see above). In summary, we can answer RQ3 and conclude that nonlinear forecasting methods outperform linear methods for car resale price forecasting when error costs are asymmetric.

Although we observe nonlinear methods to perform better than linear methods in general, it is interesting to note that the advantage of the nonlinear methods decreases with *a*; that is when the asymmetry of error costs increases. For example, we observe the overall largest and smallest difference in MQQC values between a linear and a nonlinear method at the boundary settings of *a=1.0* and *a=0.1*, respectively. Consider for example the comparison between MBL and MBNL. With MQQC=63.56, MBNL reduces error costs by almost 30 percent compared to MBL at *a=1.0*. Conversely, at *a=0.1*, MBNL performs only 1.3% better than MBL. A similar trend can be observed for other comparisons, for example between QR and QRNN. Overall, it suggests that the degree to which a method succeeds in capturing complex (nonlinear) patterns between independent variables and the target becomes less important when cost asymmetry increases. Alternatively, a high imbalance between error costs may mask complex relationships and complicate the task of nonlinear methods to discern these in a data-driven manner. Both interpretations suggest linear methods to be a viable approach for car resale price forecasting if the costs of under- versus overestimating resale prices is severely skewed. However, a detailed analysis of the economic implications associated with using a linear or nonlinear method for car resale price forecasting is beyond the scope of this study and left to future research.

5.2. Individual vs Ensemble Methods

In order to clarify the relative effectiveness of individual versus ensemble forecasts (RQ4), we perform a second set of experiments where we compare the best performing individual method, QRNN, to ensemble models developed from the base model library (Table 2). More specifically, we employ ensemble selection to identify and combine a subset of (base) model forecasts from the library using a weighted average (Caruana et al., 2006). We organize the search for the subset of models and their weights so as to minimize QQC. Lessmann (2013) provides some first, preliminary results that using ensemble selection together with ACEF works well. We further extend this approach in that we augment the model library with models that have been trained to account for asymmetric error costs. In particular, we incorporate QR, QRNN and NNAC into the model library prior to performing ensemble selection. The resulting ensemble is called ES_qqc_a, since it uses the augmented library. To increase the scope of the individual versus ensemble model comparisons, we include a second ensemble model, ES_av. which we develop as a combination of the five individual forecasting methods that address asymmetric error costs (i.e., QR, QRNN, NNAC, MBL, MBNL). Our motivation to consider this model is twofold. First, previous research suggests that the simple average is a suitable way to combine forecasts (e.g., Makridakis & Hibon, 2000). Second, Section 5.1 has demonstrated to importance to account for asymmetric error costs in the focal application. For these reasons, ES_av seems to be a suitable ensemble to consider in the analysis of RQ4.

Figure 3 3 compares the performance of ES_qqc_a and ES_av against QRNN using once again MQQC as performance measure. The abscissa represents the value of asymmetry $a = 0.1$, 0.2, …1, while $b$ is fixed at *1*. Recall that overestimating is relatively more costly than underestimating resale prices, the smaller the value of a. The y-axis shows the percentage difference between the QRNN benchmark and an ensemble. A positive difference indicates that the ensemble improves over QRNN.

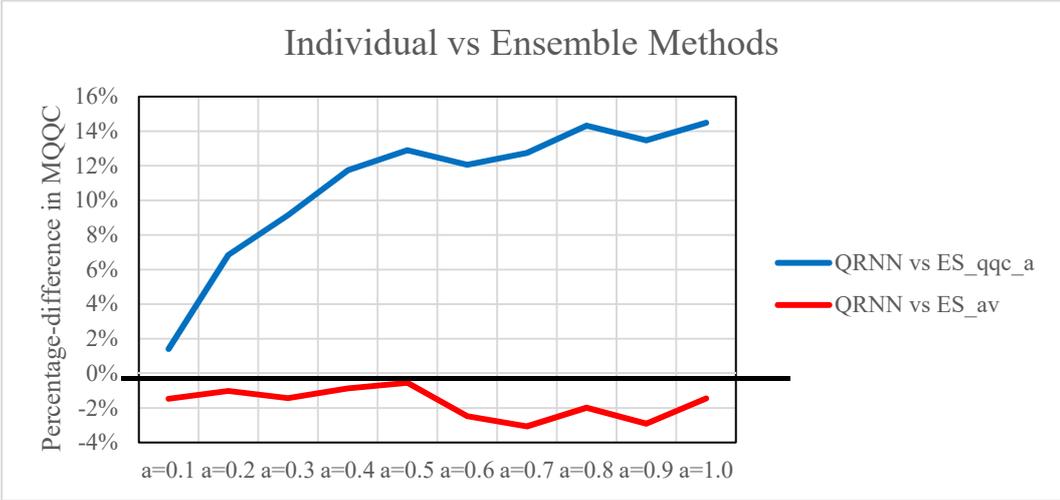

*Figure 3: Comparing the performance of the best individual model to a simple and weighted average-ensemble*

A first result of Figure 3 is that ES_av does not improve over QRNN. In fact, the performance of ES_av is less than that of QRNN for all settings of *a*. Although there is no guarantee that an ensemble improves over an individual forecasting model, much research has found ensembles to predict more accurately (e.g., Makridakis & Hibon, 2000; Timmermann, 2006). More specifically, this finding has been observed when forecasting with symmetric cost functions. In this sense, Figure 3 exemplifies that such results do not necessarily generalize to the field of forecasting with ACEF. However, a second result of Figure 3 is that a more advanced ensemble mechanism is well-capable to provide sizeable improvements over the best individual model. In particular, ES_qqc_a dominates QRNN for all settings of *a*. Last, Figure 3 also reemphasizes the trend that larger asymmetry (lower values of *a*) erodes performance differences among alternative forecasting methods.

In summary, we observe mixed results, which complicates answering our fourth research question. Ensemble forecasting does not improve over individual models in general. On the other hand, improvements over QRNN, which in view of Table 4 can be considered a powerful approach to forecast resale prices, of at least 1.8 percent and up to 14 percent under favorable conditions provide evidence that ensemble modeling is effective when forecasting with ACEF, provided that a suitable model combination strategy is employed. In particular, the ES_qqc_a approach which we propose on the basis of Caruana's et al. (2006) ensemble selection appears to be a suitable modeling strategy when the costs associated with forecast errors are asymmetric. To further substantiate this view and shed some light on the origins of the success of ES_qqc_a, we subsequently perform a sensitivity analysis.

### 5.3. Sensitivity Analysis of Ensemble Methods

The sensitivity analysis aims at gaining an understanding which factors explain the appealing performance of ES_qqc_a, which we observe in Figure 3. To that end, we examine the building blocks of ES_qqc_a and derive benchmarks that enable us to appraise the marginal utility of ensemble components. More specifically, we consider two degrees of freedom to be particularly influential. First, the model library can include symmetric models[4] only, or can be augmented with models that result from minimizing an ACEF, which we have proposed here. Second, the selection of ensemble members requires an objective, which can either be a symmetric or an asymmetric cost function. In addition, it is possible to use a symmetric cost function for ensemble selection and to bias the resulting ensemble forecasts to account for cost asymmetry (i.e., markdown approach). To understand the effect of these parameters we empirically compare all possible combinations in Table 5. The naming convention of the models is ES_{mse, md, qqc}_{c,a}, where mse, md, and qqc indicate that ensemble selection is organized so as to minimize MSE, minimize MSE with subsequent markdown adaption, and QQC, respectively, and the suffixes s and a indicate that the corresponding base model library includes symmetric models only or has been augmented with asymmetric models, respectively.

---

[4] We use the term *symmetric model* as a shorthand form for a forecasting model that has been developed through minimizing a symmetric cost function such as ordinary least squares.

*Table 5: Performance of Ensemble method measured in MQQC using all combinations of the degrees of freedom.*

| Sensitivity Analysis of Ensembles (MQQC) | | | | | | | | | | |
|---|---|---|---|---|---|---|---|---|---|---|
| Method | a=0.1 | a=0.2 | a=0.3 | a=0.4 | a=0.5 | a=0.6 | a=0.7 | a=0.8 | a=0.9 | a=1.0 |
| ES_mse_s | 34.82 | 37.27 | 39.71 | 42.15 | 44.60 | 47.04 | 49.49 | 51.93 | 54.37 | 56.82 |
| ES_md_s | 20.24 | 27.55 | 32.94 | 37.42 | 41.35 | 44.90 | 48.17 | 51.22 | 54.10 | 56.82 |
| ES_qqc_s | 22.88 | 29.03 | 33.89 | 38.05 | 41.74 | 45.15 | 48.30 | 51.29 | 54.08 | 56.82 |
| ES_mse_a | 34.45 | 36.92 | 39.38 | 41.84 | 44.31 | 46.77 | 49.23 | 51.70 | 54.16 | 56.62 |
| ES_md_a | 20.03 | 27.28 | 32.66 | 37.14 | 41.09 | 44.66 | 47.95 | 51.02 | 53.91 | 56.62 |
| ES_qqc_a | 19.68 | 26.94 | 32.40 | 36.95 | 40.98 | 44.57 | 47.86 | 50.94 | 53.87 | 56.62 |

The equivalence of MQQC values among approaches from the same library (_s or _a) in the rightmost column of Table 5 is plausible. For *a=1*, QQC reduces to MSE so that ensemble selection optimizes the same performance criterion. Also, the depreciation identified by the markdown procedure should be zero. Otherwise, ES_mse_{c,a} forecasts would have been biased beforehand.

Table 5 reveals that ES_qqc_a is the best ensemble strategy and outperforms alternative approaches for all settings of *a*. This supports our design of the ES_qqc_a ensemble. Regarding the first degree of freedom, the composition of the model library, Table 5 shows the suitability of augmenting the library with forecasting models developed from ACEF minimization. Independent of the ensemble selection criterion or the level of cost asymmetry, an augmented library consistently gives better performance than a library containing only symmetric models.

Interestingly, augmenting the model library also increases performance when costs are symmetric (rightmost column of Table 5). We attribute this effect to increased diversity among the base models if the model library includes some ACEF minimizing base models. It is well-known that increasing diversity has the potential to enhance the performance of ensemble forecasting models (e.g., Brown, et al. 2005). The idea to seek diversity through using ACEF for forecast model development has, to the best of our knowledge, not been investigated. Table 5 provides some first, preliminary evidence that such strategy might have potential. However, further research is needed to test this concept in detail.

Table 5 also clarifies the effect of the second degree of freedom, namely the selection criteria within ensemble selection. For *a=1.0*, the performance is equal for all selection criteria regardless of the type of model library. For other settings of *a*, the modeling approaches that account for cost asymmetry, ES_md_{s,a} and ES_qqc_{s,a}, consistently outperform basic ensemble selection, ES_mse_{s,a}, which is agnostic of the asymmetry among forecast error costs. However, Table 5 does not provide a clear signal as to which of the two base model selection strategies, ES_md_{s,a} or ES_qqc_{s,a}, is superior. Using a model library with symmetric models, ES_qqc_s outperforms ES_md_s at *a=0.9*. For values $0.1 \leq a \leq 0.8$, the markdown procedure gives better results. When developing ensembles from an augmented model library, ES_qqc_a outperforms ES_md_a for all settings with cost asymmetry (*a<1*).

In summary, both building blocks, the augmented model library and the selection strategy to minimize QQC contribute to the appealing performance of ES_qqc_a (e.g., in Section 5.2). Furthermore, Table 5 indicates an interaction between the two ingredients. Library augmentation is generally appropriate. Using QQC for ensemble selection, however, unfolds its full potential only if applied to a model library, which includes base models that take asymmetric error costs into account. To gain further insight into the degree to which forecasting performance increases Table 6 and Table 7 report the average percentage differences between different modeling approaches.

*Table 6: Percentage difference between ensembles from the symmetric and augmented library across the three selection criteria and settings of a. Can be interpreted as enhancement due to using the augmented library.*

| Symmetric vs. Augmented | | | | | | | | | | |
|---|---|---|---|---|---|---|---|---|---|---|
| Selection strategy | a=0.1 | a=0.2 | a=0.3 | a=0.4 | a=0.5 | a=0.6 | a=0.7 | a=0.8 | a=0.9 | a=1.0 |
| MSE | 1.06% | 0.94% | 0.83% | 0.74% | 0.65% | 0.58% | 0.51% | 0.45% | 0.39% | 0.34% |
| MD | 1.06% | 0.98% | 0.86% | 0.74% | 0.64% | 0.55% | 0.47% | 0.40% | 0.34% | 0.34% |
| QQC | 14.00% | 7.19% | 4.40% | 2.90% | 1.81% | 1.30% | 0.91% | 0.69% | 0.39% | 0.34% |

*Table 7: Percentage difference between the three types of selection criteria for the small and augmented library. Can be interpreted as enhancement by using the latter criterion compared to the first.*

| Model library | a=0.1 | a=0.2 | a=0.3 | a=0.4 | a=0.5 | a=0.6 | a=0.7 | a=0.8 | a=0.9 | a=1.0 |
|---|---|---|---|---|---|---|---|---|---|---|
| MSE vs. MD | | | | | | | | | | |
| Symmetric | 41.86% | 26.07% | 17.05% | 11.23% | 7.28% | 4.54% | 2.65% | 1.36% | 0.51% | 0.00% |
| Augmented | 41.86% | 26.09% | 17.07% | 11.23% | 7.27% | 4.52% | 2.61% | 1.31% | 0.46% | 0.00% |
| MSE vs. QQC | | | | | | | | | | |
| Symmetric | 34.30% | 22.10% | 14.65% | 9.73% | 6.42% | 4.02% | 2.40% | 1.23% | 0.55% | 0.00% |
| Augmented | 42.89% | 27.02% | 17.72% | 11.70% | 7.51% | 4.71% | 2.79% | 1.46% | 0.54% | 0.00% |
| MD vs. QQC | | | | | | | | | | |
| Symmetric | -13.01% | -5.36% | -2.89% | -1.68% | -0.93% | -0.55% | -0.26% | -0.13% | 0.03% | 0.00% |
| Augmented | 1.77% | 1.26% | 0.78% | 0.53% | 0.26% | 0.20% | 0.18% | 0.15% | 0.08% | 0.00% |

Table 6 suggests that library augmentation can achieve a small improvement in forecast performance when selecting ensemble members through MSE minimization. This is true regardless whether forecasts are adapted by means of a markdown factor. For example, even in the best setting (*a=0.1*), augmenting the model library reduces MQQC by only about one percent for MSE and MD. The corresponding improvement is as large as 14 percent (22.88 c.f. 19.68) when using QQC for ensemble selection.

Comparing Table 6 and Table 7, the specific approach to select ensemble members affects forecast performance more substantially than library augmentation. We observe the largest improvements under high degrees of cost asymmetry when moving from MSE to an alternative selection strategy, which incorporates the relative severity of different types of forecast errors. This reemphasizes the criticality to address asymmetric error costs with a suitable modeling approach and cautions against using "standard" forecasting methods (i.e., without adaption) for car resale price forecasting. In Table 6, MSE, the representative of a "standard" approach, shows the lowest performance across all settings with cost asymmetry ($a<1$).

The last panel of Table 6 compares the two forecasting approaches that account for error cost asymmetry. Interestingly, results are exactly opposite, depending on whether ensemble models are derived from the augmented model library or not. In particular, unless the library includes base model that account for asymmetric error costs, using a markdown approach is superior and decreases error costs for all settings with pronounced cost asymmetry (i.e., $a<0.9$). On the contrary, selecting ensemble members so as to minimize QQC gives better results across all settings of $a<1$ if using the augmented model library. Hence, the choice of a suitable mechanism to account for asymmetric costs of forecast errors depends on the library of base models.

In summary, the sensitivity analysis of ES_QQC_a shows that library augmentation and selection strategy are both important. They affect forecasting performance and can reduce the costs of forecast errors, respectively. The magnitude of the enhancement potential depends on the reference point. Considering, for example, the case of $a=0.5$, where negative residuals are twice as costly as positive residuals, and ES_MSE_s as reference. The latter is a sensible reference because it represents a sophisticated standard approach to develop forecasting models. Library augmentation can improve the performance of ES_MSE_c by 0.65 percent (fist row of Table 6). Using QQC instead of MSE for ensemble selection results in an enhancement of 6.42 percent (second panel of Table 7). The combination of both, as implemented in ES_QQC_a, seems to exploit synergy effects and gives an improvement of 8.11 percent (follows from Table 5, where MQQC of ES_MSE_s = 44.60 and ES_QQC_a=40.98).

Improvements of QQC over MD, if any, are much less, which could suggest that the conceptually simpler markdown approach will often by sufficient. However, consistent improvements for the overall best approach, ES_QQC_a, even if small, can be managerially meaningful in applications where a large volume of forecasts is produced.

Overall, the empirical investigation supports the view that ES_QQC_a is a suitable modeling approach when facing asymmetric costs of forecast errors. The corresponding benchmark is ensemble based on minimizing MSE, which we find to provide the most accurate forecasts in terms of MSE. Given that ES_MSE_s is built upon several base models with various combinations of meta-parameter settings, it represents a challenging benchmark. Accordingly, we find ES_MSE_s to give the lowest forecast errors in terms of MSE among all models considered in the study. Compared to this benchmark, ES_QQC_a shows high potential for further enhancement, especially if *a* is small.

## 6. Conclusion

The paper focuses on the support of pricing decisions in car leasing. Noting that informed decisions require the resale prices of return cars to be forecast and given that the cost associated with forecast errors depend on the type of error, an empirical study has been conducted to examine whether asymmetric error costs should be incorporated into forecast model development and to identify an effective approach to do so. In order to shed some light on the economic consequences of choosing one forecasting approach over another, alternative models have been assessed in terms of an ACEF with varying degrees of error cost asymmetry.

The empirical results observed in several comparisons provide strong evidence that ignoring asymmetric error costs harms decision quality. Using the QQC function as performance criterion and proxy for decision costs, we observe asymmetric forecasting methods to consistently and substantially improve upon symmetric alternatives. In several comparisons and experimental settings, we find asymmetric approach(es) to consistently reduce error costs compared to symmetric counterparts and observe cost reductions of above 40 percent over a challenging benchmark model.

Considering the question how to best account for asymmetric error costs, we compare ex post and ex ante strategies. Overall, we find evidence that an ex ante approach, which alters the development of the forecast model, is often superior to an ex post approach, which biases the forecasts of a "standard" forecasting model that does not account cost asymmetry. However, the improvement of ex ante approaches over an ex post correction is much less than the improvement over an error cost agnostic approach. Considering that an ex post adaption of forecasts is relatively easy to implement, forecasters are well advised to check carefully whether the additional effort to deploy a more sophisticated ex ante approach is well-invested. This is likely to be the case, if a large number of forecasts are generated to support business decisions at large scale. Car leasing is such an application in the sense that every single leasing contract being signed requires the lessor to forecast the resale prices of the corresponding car.

From a managerial perspective, the main implication of our study is that leasing price management should acknowledge the existence of asymmetric error costs and employ forecasting models that account that take these into account. Eventually, this helps to reduce the costs of suboptimal pricing decisions and increase efficiency in the leasing business. From an academic point of view, we close a gap in the forecasting literature through contributing empirical results related to ensemble forecasting using ACEF. Much research has shown forecast combination to improve forecast accuracy. We add to this literature by showing that previous results generalize to applications that exhibit asymmetric error costs.

The present study also suffers limitations that open up the way for future research. First, results and conclusions are restricted to the data employed here. Therefore, examining the generalizability of observed results to other domains and applications seems to be a fruitful avenue for future research. Furthermore, we observe ensemble models from augmented model libraries to perform better than models derived from original libraries of symmetric models. This finding could inspire novel methodology for ensemble forecasting with ACEF. More importantly, there might be potential to link our result to ensemble theory related to, for example, the strength-diversity-trade-off or ensemble margin.